\documentclass[letterpaper]{article}
\usepackage{aaai20}
\usepackage{times}
\usepackage{helvet}
\usepackage{courier}
\usepackage[hyphens]{url}
\usepackage{graphicx}
\usepackage{graphicx}
\usepackage{booktabs}

\title{Domain Knowledge Aided Explainable Artificial Intelligence for Intrusion Detection and Response}
\author{Sheikh Rabiul Islam, William Eberle, Sheikh K. Ghafoor, Ambareen Siraj,  Mike Rogers\\
Departmemt of Computer Science\\
Tennessee Technological University\\
Cookeville, U.S. \\
sislam42@students.tntech.edu, \{weberle, sghafoor, asiraj, mrogers\}@tntech.edu
}
\begin{document}
\maketitle
\begin{abstract}
Artificial Intelligence (AI) has become an integral part of modern-day security solutions for its ability to learn very complex functions and handling ``Big Data''. However, the lack of explainability and interpretability of successful AI models is a key stumbling block when trust in a model's prediction is critical. This leads to human intervention, which in turn results in a delayed response or decision. While there have been major advancements in the speed and performance of AI-based intrusion \textit{detection} systems, the \textit{response} is still at human speed when it comes to explaining and interpreting a specific prediction or decision. In this work, we infuse popular domain knowledge (i.e., CIA principles) in our model for better explainability and validate the approach on a network intrusion detection test case. Our experimental results suggest that the infusion of domain knowledge provides better explainability as well as a faster decision or response. In addition, the infused domain knowledge generalizes the model to work well with unknown attacks, as well as opens the path to adapt to a large stream of network traffic from numerous IoT devices.
\end{abstract}

%
%
%
%################################### INTRODUCTION ############################ 
%
%
%
\section{Introduction}
Most of the recent advancements in Artificial Intelligence (AI), and more specifically Machine Learning (ML), have come from complex non-linear models such as Deep Neural Networks, Ensemble Methods, and Support Vector Machines. These models are also known as ``black box'' models as they are complex to interpret and explain, which arises from their inherent non-linear capabilities, multiple parameters, and very complex transformations. In addition, some algorithms require a very large number of samples (i.e., large training sets) to work efficiently, where it is very difficult to figure out what the model learned from the dataset and which portion of the data set has more influence on the output \cite{interpretability}.

Due to these challenges, the black box models lacks explainability and interpretability, ultimately resulting in a lack of trust in the model and prediction, as well as possibly leading to a delayed human response/decision. This limitation also involves ethical issues in a few sensitive domains like finance (e.g., credit approval), health care (e.g., disease diagnosis), and security (e.g., identifying target). For instance, AI and ML are becoming an integral part of security solutions and defense. To mitigate the unethical use of AI as well as to promote the responsible use of AI systems, various governments have started taking different precautionary initiatives. Recently, the European Union implemented the rule of ``right of explanation'', where a user can ask an explanation of algorithmic decision \cite{goodman2017european}. In addition, more recently the US government introduced a new bill, the ``Algorithmic Accountability Act'', which would require companies to assess their machine learning systems for bias and discrimination, with a need to take corrective measures \cite{algorithmic_accountability}. The U.S. Department of Defense (DoD) has identified explainability as a key stumbling block in the adoption of AI-based solutions in many of their projects. Their DARPA division has invested \$2 billion on an Explainable Artificial Intelligence (XAI) program  \cite{explainable_ai,dark_secret_ai}.

Network intrusions are a common cyber-crime activity, estimated to cost around \$6 trillion annually in damages by 2021 \cite{false_sense_of_security}. In order to combat these attacks, an Intrusion Detection System (IDS) is a security system to monitor network and computer systems \cite{hodo2016threat}. Research in AI-based IDS has shown promising results \cite{hodo2016threat}, \cite{shone2018deep}, \cite{kim2016long}, \cite{javaid2016deep}, \cite{li2012neural}, and has become an integral part of security solutions due to its capability of learning complex, nonlinear functions and analyzing large data streams from numerous connected devices. A recent survey by \cite{dong2016comparison} suggests that deep learning-based methods are accurate and robust to a wide range of attacks and sample sizes.  However, there are concerns regarding the sustainability of current approaches (e.g., intrusion detection/prevention systems) when faced with the demands of modern networks and the increasing level of human interaction \cite{shone2018deep}. In the age of IoT and Big Data, an increasing number of connected devices and associated streams of network traffic have exacerbated the problem. In addition, the delay in detection/response increases the chance of \textit{zero day exploitation}, whereby a previously unknown vulnerability is just discovered by the attacker, and the attacker immediately initiates an attack. However, improved explainability of an AI model could quicken interpretation, making it more feasible to accelerate the \textit{response}.

Explainability is the extent to which the internal working mechanism of the machine or AI system can be explained in human terms. And interpretability is the extent to which a cause and effect (i.e., understanding of what's happening) can be observed within a system. In other words, interpretability is a form of abstract knowledge about what's happening and explainability is about the detailed step-by-step knowledge of what is happening \cite{montavon2018methods}, \cite{explainable_ai}. However, while some literature treat interpretability and explainability as the same, they are actually two different traits of a model.  Just because a model can be interpreted does not mean that it can be explained, and explainability needs to go beyond the algorithm \cite{lipton2016mythos}.     

Explainability and interpretability of a model could be achieved before, during, and after modeling. From the literature, we find that interpretability in pre-modeling (i.e., before modeling) is under-focused. \cite{miller2018explanation} argue that explainability should incorporate knowledge from a different domain such as philosophy, psychology, and cognitive science, so that the explanation is not just based on the researcher's intuition of what constitutes a good explanation. However, we also find that the use of domain knowledge for explainability is  under-focused.  In this work, we introduce a novel approach for an AI-based explainable intrusion detection and response  system, and demonstrate its effectiveness by infusing a popular network security principle (CIA principle) into the model for better explainability and interpretability of the decision.

We use a recent and comprehensive IDS dataset (CICIDS2017) which covers necessary criteria with common updated attacks such as DDoS, Brute Force, XSS, SQL Injection, Infiltration, Portscan, and Botnet. We infuse CIA principle in the model that provides a concise and interpretable set of important features.  Computer security rest on CIA principles, \textit{C} stands for \textit{confidentiality}\textemdash{concealment of information or resources},  \textit{I} stands for \textit{integrity}\textemdash{trustworthiness of data or resources}, and \textit{A} stands for \textit{availability}\textemdash{ability to use the information or resource desired}  \cite{matt2006introduction}. For instance, security compromise in \textit{confidentiality} could happen through eavesdropping unencrypted data,  compromise in \textit{integrity} could happen through an unauthorized attempt to change data, and the compromise in \textit{availability} could happen through the deliberate arrangement of denial to access data or service.

We also convert the domain knowledge infused features into three features C, I, and A by quantitatively computing compromises associated with each of those for each record. Output expressed as these generalized and newly constructed set of features provides better explainability with negligible compromises in performance.  We also found that generalization provides more resiliency against unknown attacks.   

In summary, our contributions in this work are as follows: (1) we demonstrate a method for the collection and use of domain knowledge in an intrusion detection/response system; (2) we introduce a way to bring popular security principles (e.g., CIA principles)  to aid in interpretability and explainability;  (3) our experimental results show that infusing domain knowledge into ``black box'' models can make them better explainable with little or no compromise in performance ; and (4) domain knowledge infusion increases generalizability, which leads to better resiliency against unknown attack. 

We start with a background of related work (Section \ref{sec:background}) followed by a description of our proposed approach, an intuitive description of standard supervised algorithms, and an overview of the
dataset (Section \ref{sec:methodology}) used in this work. In Section \ref{sec:experiment}, we describe our experiments, followed by Section \ref{sec:results} which contains discussion on results from the experiments. We conclude with limitations and future work in Section \ref{sec:conclusion}.
%
%
%
%###################################  BACKGROUND ############################ 
%
%
%
\section{Background} \label{sec:background}
Research in Explainable Artificial Intelligence (XAI) is a re-emerging field, after the earlier work of  \cite{chandrasekaran1989explaining}, \cite{swartout1993explanation}, and  \cite{swartout1985rule}. Previous work focused on primarily explaining the decision process of knowledge-based systems and expert systems.  The classical learning paradigm Explanation-Based Learning (EBL), introduced in the early '80s, can also be regarded as a precursor of explainability. EBL involves learning a problem-solving technique by observing and analyzing solutions to a specific problem \cite{dejong1981generalizations}, \cite{mitchell1986explanation}. The main reason for the renewed interest in XAI research has stemmed from recent advancements in AI and ML and their application to a wide range of areas, as well as concerns over unethical use and undesired biases in the models. In addition, recent concerns and laws by different governments are necessitating more research in XAI. 

\cite{yang2017explainable}, use \textit{Bayesian Teaching}, where a smaller subset of examples is used to train the model instead of the whole dataset. The subset of examples is chosen by domain experts as the examples are most relevant to the problem of interest. However, for this purpose, choosing the right subset of examples in the real-world is challenging.

\cite{lei2016rationalizing} propose an approach for sentiment analysis where a subset of text from the whole text is selected as the rationale for the prediction. In addition, the selected subset of text is concise and sufficient enough to act as a substitute for the original text, and still capable of making the correct prediction.  Although their approach outperforms available attention-based models (from deep learning) with variable-length input (e.g., a model for document summarization) , it is limited to only text analysis. 

When the explanation is based on feature importance, it is necessary to keep in mind that features that are globally important may not be important in the local context, and vice versa \cite{ribeiro2016should}. \cite{ribeiro2016should} propose a novel explanation technique capable of explaining the prediction of any classifier (i.e., in the model agnostic way) with a locally interpretable model (i.e., in the vicinity of the instance being predicted) around the prediction. Their concern is on two issues: (1) whether the user should trust the prediction of the model and act on that, and (2) whether the user should trust a model to behave reasonably-well when deployed. In addition, they involve human judgment in their experiment (i.e., human in the loop) to decide whether to trust the model or not.

\cite{kim2017interpretability} propose a concept attribution-based approach (i.e., sensitivity to concept) that provides an interpretation of the neural network's internal state in terms of human-friendly concepts. Their approach, \textit{Testing with CAV (TCAV)}, quantifies the prediction's sensitivity to the high dimensional concept. For example, a user-defined set of examples that defines the concept ``striped'', TCAV can quantify the influence of  ``striped'' in the prediction of ``zebra'' as a single number. To learn the high dimensional concepts they use a Concept Activation Vector (CAV) \textemdash{CAVs are learned from training a linear classifier that can distinguish between the activations produced by a particular concept's examples and examples in any layer}. 

Most of these approaches try to find out how the prediction deviates from the base/average scenario. Lime \cite{ribeiro2016should} tries to generate an explanation by locally (i.e., using local behavior) approximating the model with an interpretable model (e.g., decision trees, linear model). However, it is limited by the use of the only linear model to approximate the local behavior. \cite{lundberg2017unified} propose  ``SHAP'' which unifies seven previous approaches: LIME \cite{ribeiro2016should}, DeepLIFT \cite{shrikumar2017learning}, Tree Interpreter \cite{on_tree_interpreter}, QII \cite{datta2016algorithmic}, Shapley sampling values \cite{vstrumbelj2014explaining}, Shapley regression values \cite{lipovetsky2001analysis}, and Layer-wise relevance propagation \cite{bach2015pixel} to make the explanation of prediction for any machine learning model. While SHAP comes with theoretical guarantees about consistency and local accuracy from game theory, it needs to run many evaluations of the original model to estimate a single vector of feature importance \cite{shap_vs_lime}. ELI5 also uses the LIME algorithm internally for explanations. In addition, ELI5 is not truly model agnostic, mostly limited to tree-based and other parametric or linear models. Furthermore, Tree Interpreter is limited to only tree-based approaches (e.g., Random Forest, Decision Trees). 

AI-based IDSs have continued to show promising performance \cite{hodo2016threat},\cite{shone2018deep},\cite{kim2016long},\cite{javaid2016deep},\cite{li2012neural}. \cite{shone2018deep} propose an approach in combination of both shallow (Random Forest) and deep learning (Auto Encoder), capable of analyzing a wide range of network traffic, outperforming mainstream Deep Belief Networks (DBN). In a literature survey on traditional IDS vs deep learning IDS by \cite{dong2016comparison}, they suggest deep learning-based methods provide better accuracy for a wide range of samples sizes and a variety of network traffic or attacks \cite{dong2016comparison}. However, in all of the previous work, there are still long training times and a reliance on a human operator \cite{shone2018deep}. 

However, incorporating domain knowledge for explainability has garnered little attention. Previously, we introduced the concept of infusing domain knowledge \cite{islam2019infusing}, albeit for bankruptcy prediction with a limited focus. \cite{miller2018explanation} have argued that incorporating knowledge from different domains will provide better explainability. In addition, \cite{kim2017interpretability} use the prediction's sensitivity to high dimensional concepts (e.g., the concept ``striped'' to ``Zebra'') for explaining the prediction. Furthermore, both LIME \cite{ribeiro2016should} and SHAP \cite{lundberg2017unified} use a simplified input mapping\textemdash{mapping the original input to a simplified set of input}. To the best of our knowledge, none of the models incorporate domain knowledge with a focus towards better explainability and interpretability. Although our proposed conceptual model comes with a negligible compromise in accuracy, it comes with better explainability and interpretibility, and scalability to big data problems.

%
%
%
%###################################  METHODOLOGY ############################ 
%
%
%
\section{Methodology} \label{sec:methodology}

\begin{figure}[h]
  \centering
  \includegraphics[width=\linewidth]{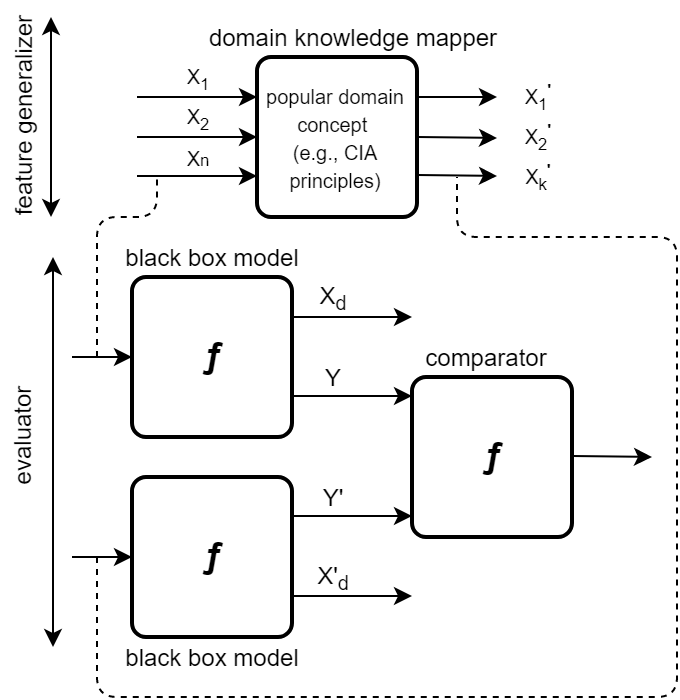}
  \caption{Proposed Technique}
    \label{fig:proposed_technique}
\end{figure}

\subsection{Proposed Approach}\label{subsec:proposed_approach}

The proposed approach consists of two components: a \textit{feature generalizer}, which gives a generalized feature set with the help of domain knowledge in two different ways; and an \textit{evaluator} that produces and compares the results from the ``black box'' model for multiple configurations of features: domain knowledge infused features, newly constructed features from domain knowledge infused features, selected features, and all features. 

\subsection{Feature Generalizer}\label{subsec:feature_generalizer}

\begin{table}
\caption{Mapping of network attack with related component of CIA principles}
\label{tab:mapping_attack_vs_cia}
\centering
\begin{tabular}{ll}
\toprule
Attack        & Related component of CIA  \\
\midrule
DoS GoldenEye & A                   \\
Heartbleed    & C                   \\
DoS hulk      & A                   \\
DoS Slowhttp  & A                   \\
DoS slowloris & A                   \\
SSH-Patator   & C                   \\
FTP-Patator   & C                   \\
Web Attack    & C, I, A                 \\
Inflitration  & C                   \\
Bot           & C, I, A                 \\
PortScan      & C                   \\
\textbf{DDoS}          & \textbf{A}                   \\
\bottomrule
\end{tabular}
\end{table}

\begin{table*}
\tiny
\caption{Mapping of feature with related component of CIA principles}
\label{tab:mapping_feature_vs_cia}
\centering
\begin{tabular}{lp{6cm}p{3cm}l}
\toprule
Feature                     & Description                                                                                     & In top 3 features of attack                                           & Renamed feature                    \\
\midrule
ACK Flag Count              & Number of packets with ACK                                                                      & SSH-Patator                                                           & ACK Flag Count - C                 \\
Active Mean                 & Mean time a flow was active before becoming idle                                                & DoS Slowhttp, Infiltration                                            & Active Mean - AC                   \\
Active Min                  & Minimum time a flow was active before becoming idle                                             & DoS Slowhttp                                                          & Active Min - A                     \\
\textbf{Average Packet Size}         & \textbf{Average size of packet}                                                                          & \textbf{DDoS}                                                                  & \textbf{Avg Packet Size - A}                \\
Bwd IAT Mean                & Mean time between two packets sent in the backward direction                                    & DoS slowloris                                                         & Bwd IAT Mean - A                   \\
Bwd Packet Length Std       & Standard deviation size of packet in backward direction                                         & DoS Hulk, DoS GoldenEye, DDoS, Heartbleed, DoS Hulk                   & Bwd Packet Length Std - AC         \\
Bwd Packets/s               & Number of backward packets per second                                                           & Bot, PortScan                                                         & Bwd Packets/s - CIA                \\
Fwd IAT Mean                & Mean time between two packets sent in the forward direction                                     & DoS slowloris                                                         & Fwd IAT Mean - A                   \\
Fwd IAT Min                 & Minimum time between two packets sent in the forward direction                                  & DoS slowloris, DoS GoldenEye                                          & Fwd IAT Min - A                    \\
Fwd Packet Length Mean      & Mean size of packet in forward direction                                                        & Benign, Bot                                                           & Fwd Packet Length Mean - CIA       \\
Fwd Packets/s               & Number of forward packets per second                                                            & FTP-Patator                                                           & Fwd Packets/s - C                  \\
Fwd PSH Flags               & Number of times the PSH flag was set in packets travelling in the forward direction (0 for UDP) & FTP-Patator                                                           & Fwd PSH Flags - C                  \\
Flow Duration               & Duration of the flow in Microsecond                                                             & DDoS, DoS slowloris, DoS Hulk, DoS Slowhttp, Infiltration, Heartbleed & Flow Duration - AC                 \\
Flow IAT Mean               & Mean inter-arrival time of packet                                                               & DoS GoldenEye                                                         & Flow IAT Mean - A                  \\
Flow IAT Min                & Minimum inter-arrival time of packet                                                            & DoS GoldenEye                                                         & Flow IAT Min - A                   \\
Flow IAT Std                & Standard deviation inter-arrival time of packet                                                 & DDoS, DoS Slowhttp, DoS Hulk                                          & Flow IAT Std - A                   \\
Init\_Win\_bytes\_forward   & The total number of bytes sent in initial window in the backward direction                      & Web Attack                                                            & Init Win Bytes Fwd - CIA           \\
PSH Flag Count              & Number of packets with PUSH                                                                     & PortScan                                                              & PSH Flag Count - C                 \\
Subflow Fwd Bytes           & The average number of packets in a sub flow in the forward direction                            & Benign, SSH-Patator, Web Attack, Bot, Heartbleed, Infiltration        & Subflow Fwd Bytes - CIA            \\
SYN Flag Count              & Number of packets with SYN                                                                      & FTP-Patator                                                           & SYN Flag Count - C                 \\
Total Length of Fwd Packets & Total size of packet in forward direction                                                       & Benign, SSH-Patator, Web Attack, Bot, Heartbleed, Infiltration        & Total Length of Fwd Packets - CIA \\
\bottomrule
\end{tabular}
\end{table*}

The \textit{feature generalizer} (Figure \ref{fig:proposed_technique}, top portion), takes original features of the dataset (\textit{X}\textsubscript{1}, \textit{X}\textsubscript{2}, .... \textit{X}\textsubscript{n} \begin{math} \in \end{math} \textit{X} where \textit{X} is the set of all features) and infuse domain knowledge to produce/re-construct a concise and better interpretable feature set (\textit{X}\textsubscript{1}', \textit{X}\textsubscript{2}', ..... \textit{X}\textsubscript{k}'  \begin{math} \in \end{math} \textit{X'} where \textit{X'} is the universal set of original/transformed/constructed features, but here \textit{k} is much smaller than \textit{n}) in two different ways:
\subsubsection{Feature Mapping}\label{subsec:feature_mapping}
As stated earlier, we use CIA principles as domain knowledge, which stands for \textit{confidentiality}, \textit{integrity}, and \textit{availability}. We analyze all types of attacks for associated compromises in each component of CIA principles (see Table \ref{tab:mapping_attack_vs_cia}). The \textit{Heartbleed} vulnerability is related to a compromise in \textit{confidentiality} as an attacker could gain access to the memory of the systems protected by the vulnerable version of the OpenSSL. A \textit{Web attack} (e.g., Sql injection) is related to a compromise in \textit{confidentiality} and \textit{integrity} (e.g., read/write data using injected query), and \textit{availability} (e.g., flooding the database server with injected complex queries like a cross join). \textit{Infiltration} attack is related to a compromise in \textit{confidentiality} as it normally exploits software vulnerability (e.g., Adobe Acrobat Reader) to create a backdoor and reveal information (e.g., IP’s). \textit{Port scan} attack is related to a compromise in \textit{confidentiality} as the attacker sends packets with varying destination ports to learn the services and operating systems from the reply. All \textit{DoS} and \textit{DDoS} attacks are related to a compromise in \textit{availability} as it aims to hamper the availability of service or data. Furthermore, \textit{SSH patator} and \textit{FTP patator} are brute force attacks and are usually responsible for a compromise in \textit{confidentiality}. Botnet (i.e., robot network\textemdash{}a network of malware-infected computers) could provide a remote shell, file upload/download option, screenshot capture option, and key logging options which has potential for all of the \textit{confidentiality, integrity, and availability} related compromises.

Furthermore, from the feature ranking of the original dataset provider \cite{sharafaldin2018toward}, for each type of attack, we take the top three features according to their importance (i.e., feature importance from Random Forest Regressor) and calculate the mapping (Table \ref{tab:mapping_feature_vs_cia}) with related compromises under CIA principles. For example, the feature \textit{Average Packet Size} is renamed as \textit{Avg Packet Size - A} where -A indicates that it is a key feature for the compromise of \textit{availability} (see Table \ref{tab:mapping_feature_vs_cia}). To get this mapping between feature and associated compromises, we first find the mapping between an attack and related compromises (from Table \ref{tab:mapping_attack_vs_cia}, formulated as Equation \ref{eq:formula2}). In other words, Formula \ref{eq:formula1} gives the name of the associated attack where the feature is in the top three feature to identify that particular attack and Formula \ref{eq:formula2} gives associated compromises in C, I, or A from the attack name. Thus, with the help of domain knowledge, we keep 22 features (see Table \ref{tab:mapping_feature_vs_cia}) out of a total of 78 features.  We will refer to these features as the \textit{domain features}. The feature descriptions in Table \ref{tab:mapping_feature_vs_cia} are taken from the data processing software's website \cite{netflowmeter}.
\begin{equation} \label{eq:formula1}
f(feature)\rightarrow attack 
\end{equation}
\begin{equation} \label{eq:formula2}
f(attack)\rightarrow C, I, or A
\end{equation}
\subsubsection{Feature Construction}\label{subsec:feature_construction}
We also construct three new features, C, I, and A, from the domain features by quantitatively calculating compromises associated with each of the domain features. For that purpose, we calculate the correlation coefficient vector of the dataset to understand whether the increase in the value of a feature has a positive or negative impact on the target variable. We then convert the correlation coefficient (a.k.a \textit{coeff}) vector V in to a 1 or -1 based on whether the correlation coefficient is positive or negative accordingly. We also group the domain features and corresponding \textit{coeff} tuple into three groups. Using formula \ref{eq:formula3}, \ref{eq:formula4}, and \ref{eq:formula5}, we aggregate each group  (from C, I, and A) of domain features into the three new features C, I, and A. We also scale all feature values from 0 to 1 before starting the aggregation process. During the aggregation for a particular group (e.g., C), if the correlation coefficient vector (e.g., V\textsubscript{i}) for a feature (e.g., C\textsubscript{i}) of that group has a negative value, then the product of the feature value and the correlation coefficient for that feature is deducted, and vice-versa if positive. In addition, when a feature is liable for more than one compromise, the feature value is split between the associated elements of CIA principles. 
\begin{equation} \label{eq:formula3}
 C = \sum_{i=0}^{n} C\textsubscript{i} V\textsubscript{i}
\end{equation}
\begin{equation} \label{eq:formula4}
 I = \sum_{i=0}^{n} I\textsubscript{i} V\textsubscript{i}
\end{equation}
\begin{equation} \label{eq:formula5}
 A = \sum_{i=0}^{n} A\textsubscript{i} V\textsubscript{i}
\end{equation}
\subsection{Evaluator}\label{subsec:evaluator}
The task of the \textit{evaluator} (Figure \ref{fig:proposed_technique}, bottom side) is to execute (supervised models or algorithms) and compare the performance (in detecting malicious and benign records) of four different types of configurations of features, as follows: (1) using all features, (2) using selected features (selection is done by feature selection algorithm), (3) using domain knowledge infused features, and (4) using newly constructed features C, I, and A from domain knowledge infused features.  In addition, the \textit{evaluator} performs the following two tests:
\begin{enumerate}
    \item Explainability Test: The purpose of this test is to discover the comparative advantages or disadvantages of incorporating domain knowledge in the experiment; and
    \item Generalizability Test: The purpose of this test is to analyze how different approaches perform in unknown or unseen attack detection.  We delete all training records for a particular attack one at a time and investigate the performance of the model on the same test set, which includes records from unknown or unseen attacks. Details of these tests are described in Section \ref{sec:experiment}. 
\end{enumerate}

\subsection{Algorithms}\label{subsec:algorithms}
We use six different algorithms for predicting malicious records: one of those is a probabilistic classifier based on Naive Bayes theorem, and the remaining five are supervised ``black box'' models. The algorithm descriptions are taken from our previous work \cite{islam2019infusing}.

\subsubsection{Artificial Neural Network (ANN)}
An Artificial Neural Network is a non-linear model, capable of mimicking human brain
functions to some extent. It consists of an input layer, one or multiple hidden layer(s), and the output layer. Each layer
consists of multiple neurons that help to learn the complex pattern. Each subsequent layer learns more abstract concepts before it finally merges into the output layer.

\subsubsection{Support Vector Machine (SVM)}
The Support Vector Machine (SVM) was first introduced by  \cite{boser1992training} and has been used for many supervised classification tasks. In addition to linear classification, the model can learn an optimal
hyperplane that separates instances of different classes using a highly non-linear implicit mapping of input
vectors in high dimensional feature space (i.e., kernel trick) \cite{hooman2016statistical}. When the number of samples is too high (i.e., millions) then it is very costly in terms of computation time.

\subsubsection{Random Forest (RF)}
A Random Forest is a tree-based ensemble technique developed by \cite{breiman2001random} for the supervised classification task. In RF, many trees are generated from the bootstrapped subsamples (i.e., random sample drawn with replacement) of the training data. In each tree, the splitting attribute is chosen from a smaller random subset of attributes of that tree (i.e., the chosen split attribute that is the best among that random subset). This randomness helps to make trees less correlated as correlated trees make the same kinds of prediction errors and can overfit the model. In less correlated trees, a few trees may be wrong but many others will be right and as a group the trees can move in the right direction as the output from all the trees are averaged for the final prediction. 

\subsubsection{Extra Trees (ET)}
Extremely Randomized Trees or Extra Trees (ET) is a tree-based ensemble
technique simialr to  RF. The only difference is in
the process of splitting attribute selections and determining the threshold (cutoff) value, both are chosen in an extremely random fashion \cite{islam2018mining}. Similar to RF, a random subset of
features are taken into consideration for the split selection, but instead of choosing the most discriminative cut off threshold, ET cut off thresholds are set to random values. Thus, the best of these randomly chosen values is set as the threshold for the splitting rule \cite{ensemble_methods} on a particular node. Unlike DT, RF has multiple trees which leads to a reduced variance. However, bias is introduced, as a subset of the whole feature set is chosen for each tree instead of all features. ET was proposed by \cite{geurts2006extremely}, and has achieved a state of the art performance in some anomaly/intrusion detection research \cite{islam2018efficient}, \cite{islam2018credit},\cite{islam2018mining}.

\subsubsection{Gradient Boosting (GB)}
\cite{friedman2001greedy}, generalized Adaboost to a Gradient Boosting algorithm to allow a variety of loss function. Here the shortcoming of weak learners is identified using the gradient instead of highly weighted data points as in Adaboost. Gradient Boosting (GB) is a
classifier/regression model in the form of an ensemble of weak prediction models, such as  Decision Trees. It works sequentially like the 
AdaBoost algorithm, in that each subsequent model tries to minimize the loss function (i.e., Mean Squared Error) by paying special focus on instances that were hard to get right in the previous model.

\subsubsection{Naive Bayes (NB)}
Naive Bayes algorithm is based on Bayes Theorem, which was formulated in the seventeenth century. It is a supervised, simple, and comparatively fast algorithm based on statistics. In a real-world problem, it is unusual that all features are independent. However, Naive Bayes assumes conditional independence among features and surprisingly works well in many cases. It also requires a small amount of training data to estimate the necessary parameters \cite{naive_bayes}. This assumption of Naive Bayes helps to avoid lots of computations (e.g., computing the conditional probability for each feature with others) and makes it a faster algorithm. Besides, the avoidance of a conditional probability calculation helps (the class conditional feature distribution can be independently estimated as one-dimensional distribution) in Big Data problems where the curse of dimensionality is a concern.  However, NB is a bad estimator of a probabilty \cite{zhang2004optimality}. We use the Bernoulli Naive Bayes \cite{manning2010introduction} for our experiments where each feature is assumed to be binary-valued.

\subsection{Data}\label{subsec:data}
In this work, we use a recent and comprehensive IDS dataset namely CICIDS2017, published in 2017, covers necessary criteria with common updated attacks such as DoS, DDoS, Brute Force, XSS, SQL Injection, Infiltration, Portscan, and Botnet. In fact, this dataset is created to eliminates the shortcomings (e.g., lack of traffic diversity and volume, lack of variety of attacks, anonymized packet information, and out of date)  of previous well known IDS dataset such as DARPA98, KDD99, ISC2012, ADFA13, DEFCON, CAIDA, LBNL, CDX, Kyoto, Twente, and UMASS since 1998.  This is a labeled dataset containing 78 network traffic features (some features are listed in Table \ref{tab:mapping_feature_vs_cia}) extracted and calculated from pcap file using CICFlowMeter software \cite{lashkari2017characterization} for all benign and intrusive flows \cite{sharafaldin2018toward} . This new IDS dataset includes seven common updated family of attacks satisfying real-world criteria, also publicly available at here: https://www.unb.ca/cic/datasets/ids-2017.html .

Each record of the dataset is labeled by the particular type of attack. We make a new feature ``Class'', which is the target feature. We set the value of the ``Class'' attribute to 1 for all records labeled as any of 14 types of attacks, as those are malicious/intrusive, and set the value to 0 for the remaining records as those are benign. Following that, in the whole dataset,  there are total 2,830,743 records for 14 different attacks, 2,273,097 are benign and 557,646 are malicious. Approximately 24.5\% of the records are malicious, giving us an imbalanced dataset which impacts the performance (e.g., bias to the class of majority samples) of some machine learning algorithms. To overcome this problem, we use the well-known oversampling technique SMOTE \cite{chawla2002smote} to oversample the minority class. In \cite{dong2016comparison}, the author uses SMOTE to overcome the issue in their empirical study on the comparison of traditional vs deep learning-based IDS. SMOTE creates synthetic samples rather than just oversampling with replacement. The minority class is oversampled by creating new examples along with the line segments joining any or all of k nearest minority samples, where k is chosen based on the percentage of oversampling required (i.e., hyperparameter to the algorithm) \cite{chawla2002smote}.
%
%
%
%###################################  EXPERIMENT ############################ 
%
%
%
\section{Experiments}\label{sec:experiment}

\subsection{Experimental Setup}\label{subsec:experimental_setup}
We execute the experiments in a GPU enabled Linux machine with 12GB of RAM and core i7 processor. All supervised machine learning algorithms are implemented using the Python-based \textit{Scikit-learn} \cite{scikit-learn} library. In addition, we use \textit{Tensorflow} \cite{tensorflow} for the Artificial Neural Network. 
Due to resource limitations, instead of using the whole dataset, we take a stratified sample of the data which is big enough (i.e., 300K records) for a single GPU enabled commodity machine. We make the sampled dataset available to the research community at  \cite{sampled_dataset}. Furthermore, we use 70\% of the data for training the models and kept 30\% of the data as a holdout set to test the model. We confirm the target class had the same ratio in both sets. To avoid the adverse effect of class imbalance in classification performance, we re-sample the minority class of the training set using SMOTE \cite{chawla2002smote} to balance the dataset. However, we do not re-sample the test set, as real-world data is skewed and oversampling the test set could exhibit an overoptimistic performance. 

We run all supervised machine learning algorithm using four different approaches: 
\begin{enumerate}
\item With all features: using all 78 features of the dataset without discarding any features.
\item With selected features: using \textit{Random Forest Regressor} (adapting with the work of \cite{sharafaldin2018toward}) to select important features of the dataset, giving us 50 important features having a nonzero influence on the target variable;   
\item With domain knowledge infused features: using infused domain knowledge features (see Section \ref{subsec:feature_mapping}), we will use the term \textit{domain features} interchangeably to express it in short form; and  
\item With newly constructed features from domain knowledge infused features: using newly constructed features C, I, and A (see Section \ref{subsec:feature_construction}) from domain knowledge infused features, we will use the term \textit{domain features-constructed} interchangeably to express it in short form.  
\end{enumerate}

The following are two types of experiments using each of the four feature settings.

\subsection{Explainability Test}\label{subsec:explainability_test}
For this test, we run six supervised algorithms RF, ET, SVM, GB, ANN, and NB using the four described feature settings and report the results Section \ref{subsec:result_explainability}. Unlike NB, other classifiers are ``black box'' in nature.  NB is a probabilistic classifier based on Bayes Theorem with strong conditional independence assumption among features. The main purpose to include NB in the experiment is the generalizability test.

\subsection{Generalizability Test}\label{subsec:generalizability_test}
For testing the generalizability of the approach, we train the classifier without the representative of a particular attack, and test it with the presence of the representative of that particular attack, in order to classify it malicious/benign. To be more specific, we delete all records of a particular attack from the training set, train the classifier with the records of the remaining 13 attacks, and test the classifier with all 14 attacks. We report the percentage of deleted attacks that are correctly detected as malicious (see Section \ref{subsec:res_generalizability}). We repeat this one by one for all 14 attacks. We make the source code available to the research community to replicate the experiments at \cite{project_code2}.  
   
%
%
%
%###################################  RESULTS  ############################ 
%
%
%
 \section{Results}\label{sec:results}
The following sections discuss results from the two categories of experiments previously described. 

\subsection{Findings from Explainability Test}\label{subsec:result_explainability}
Comparing the performance using \textit{all features} vs \textit{selected features}, Table \ref{tab:all_vs_sel} shows that models using \textit{all features} (denoted with an appended -A, for instance RF-A) tend to show better results in terms of all performance metrics. However, while the difference with the \textit{selected features} setting is negligible (\textless .0007 for RF) for any performance metric, that might be a result of the elimination of features with little significance. In addition, Random Forest outperforms other algorithms SVM, ET, GB, ANN, and NB under this feature setting (i.e., using all features). So we consider the results using all features as a baseline to compare against our proposed approach. 
\begin{table}
\tiny
\centering
\caption{Performance using all features vs selected features}
\label{tab:all_vs_sel}
\begin{tabular}{llllll}
\toprule
Alg.  & Acc.   & Prec.  & Rec.     & F-score & AUC       \\
\midrule
RF-A  & \textbf{0.9987} & \textbf{0.9965} & \textbf{0.9971}   & \textbf{0.9968}  & \textbf{0.9997}    \\
RF-S  & 0.9986 & 0.9962 & 0.9966   & 0.9964  & \textbf{0.9997}    \\
\textit{Difference} & 0.0002 & 0.0003 & 0.0006   & 0.0005  & 0.0000    \\
ET-A  & 0.9981 & 0.9951 & 0.9951   & 0.9951  & 0.9994    \\
ET-S  & 0.9980 & 0.9950 & 0.9950   & 0.9950  & 0.9994    \\
\textit{Difference} & 0.0001 & 0.0002 & 0.0001   & 0.0001  & 0.0000    \\
ANN-A & 0.9802 & 0.9155 & 0.9908   & 0.9516  & 0.9984    \\
ANN-S & 0.9740 & 0.8929 & 0.9860   & 0.9372  & 0.9968    \\
\textit{Difference} & 0.0062 & 0.0226 & 0.0047   & 0.0145  & 0.0017    \\
SVM-A & 0.9109 & 0.6996 & 0.9595   & 0.8092  & 0.9780    \\
SVM-S & 0.8869 & 0.6433 & 0.9565   & 0.7692  & 0.9746    \\
\textit{Difference} & 0.0239 & 0.0563 & 0.0030   & 0.0400  & 0.0034    \\
GB-A  & 0.9960 & 0.9854 & 0.9944   & 0.9899  & 0.9995    \\
GB-S  & 0.9957 & 0.9840 & 0.9945   & 0.9892  & 0.9996    \\
\textit{Difference} & 0.0003 & 0.0014 & (0.0001) & 0.0007  & (0.0001)  \\
NB-A  & 0.7753 & 0.4371 & 0.4888   & 0.4615  & 0.8601    \\
NB-S  & 0.7621 & 0.4144 & 0.5019   & 0.4539  & 0.8508    \\
\textit{Difference} & 0.0132 & 0.0228 & (0.0131) & 0.0076  & 0.0093    \\
\bottomrule
\end{tabular}
\end{table}

Before starting the comparison of results from our approach with all features (i.e., baseline), we seek the best feature setting among two domain related feature settings of our proposed approach. In other words, in our attempt to find the better approach among using domain knowledge infused features vs newly constructed features (C, I, and A) from domain knowledge infused features, we find that, in almost all cases, the model with domain knowledge infused features (denoted with an appended -D1, for instance RF-D1) performs better than the counterpart (see Table \ref{tab:dom_vs_con}). Although for RF, the maximum performance gap is .2 in the recall, for ET that gap is .048 with a similar precision. As the domain features (22 features) contain a lot more detail than the newly constructed features C, I, and A (3 features), it loses few details. In terms of individual algorithms, RF is again a clear winner this time using domain features. Although NB and ANN exhibit better recall using constructed features, it comes with compromises in precision. So, overall we consider the domain features setting as the best over the constructed features. 
\begin{table}
\tiny
\centering
\caption{Performance using domain features vs constructed features}
\label{tab:dom_vs_con}
\begin{tabular}{llllll}
\toprule
Alg.   & Acc.   & Prec.  & Rec.     & F-score & AUC     \\
\midrule
RF-D1  & \textbf{0.9973} & \textbf{0.9920} & \textbf{0.9945}   & \textbf{0.9932}  & \textbf{0.9993}  \\
RF-D2  & 0.9511 & 0.9446 & 0.7985   & 0.8654  & 0.9572  \\
\textit{Difference} & 0.0463 & 0.0475 & 0.1960   & 0.1278  & 0.0421  \\
ET-D1  & 0.9969 & 0.9913 & 0.9932   & 0.9923  & 0.9989  \\
ET-D2  & 0.9756 & 0.9321 & 0.9448   & 0.9384  & 0.9954  \\
\textit{Difference} & 0.0214 & 0.0592 & 0.0483   & 0.0538  & 0.0036  \\
ANN-D1 & 0.9497 & 0.8300 & 0.9362   & 0.8799  & 0.9865  \\
ANN-D2 & 0.5952 & 0.3241 & 0.9721   & 0.4862  & 0.7921  \\
\textit{Difference} & 0.3544 & 0.5059 & (0.0359) & 0.3937  & 0.1945  \\
SVM-D1 & 0.8489 & 0.5747 & 0.8968   & 0.7005  & 0.9252  \\
SVM-D2 & 0.7195 & 0.3739 & 0.6281   & 0.4687  & 0.7886  \\
\textit{Difference} & 0.1294 & 0.2008 & 0.2687   & 0.2318  & 0.1366  \\
GB-D1  & 0.9881 & 0.9513 & 0.9904   & 0.9705  & 0.9986  \\
GB-D2  & 0.9230 & 0.7692 & 0.8701   & 0.8165  & 0.9789  \\
\textit{Difference} & 0.0652 & 0.1821 & 0.1204   & 0.1539  & 0.0198  \\
NB-D1  & 0.7982 & 0.4881 & 0.5028   & 0.4953  & 0.8553  \\
NB-D2  & 0.5591 & 0.2687 & 0.7195   & 0.3913  & 0.6591  \\
\textit{Difference} & 0.2391 & 0.2194 & (0.2167) & 0.1040  & 0.1962  \\
\bottomrule
\end{tabular}
\end{table}

\begin{figure}[h]
  \centering
  \includegraphics[width=\linewidth]{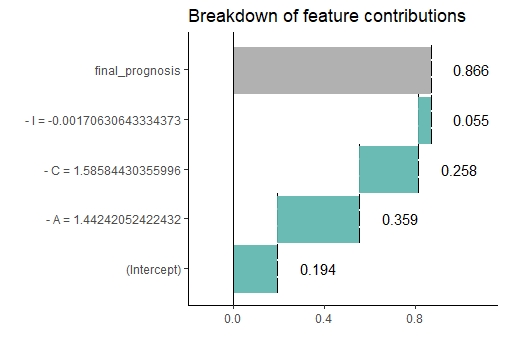}
  \caption{Breakdown of the prediction for a random sample in terms of C, I, and A.}
    \label{fig:breakdown}
\end{figure}

\begin{figure}[h]
  \centering
  \includegraphics[width=\linewidth]{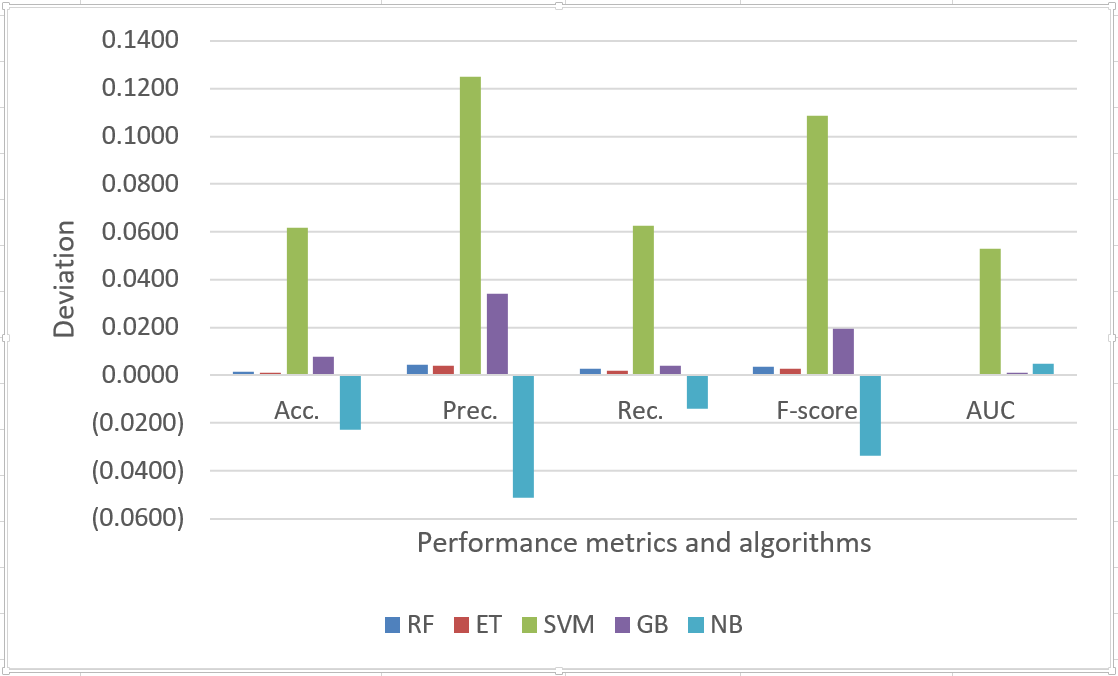}
  \caption{Performance deviations of using domain knowledge infused features from baseline}
    \label{fig:all_vs_dm}
\end{figure}

While we know the best feature setting is the \textit{all features}, as shown in the comparison of \textit{all features} vs \textit{selected features} in the Table \ref{tab:all_vs_sel}), we also know the best feature setting \textit{domain features} from \textit{domain features} vs \textit{constructed features} (see Table  \ref{tab:dom_vs_con}). So we further compare the performance of models using  the two best settings \textit{all features} (i.e., baseline) vs \textit{domain features}. We find that, among all models, RF using all features (denoted with an appended -A, for instance RF-A) performs better than all other algorithms (see Table \ref{tab:all_vs_dom} and Figure \ref{fig:all_vs_dm}). Interestingly, RF using domain knowledge infused features (denoted with an appended -D1, for instance RF-D1) also shows promising performance. The difference between these two in terms of any performance metrics is negligible (\textless .005). In fact, the result of RF using the domain knowledge infused feature settings is better than what \cite{sharafaldin2018toward} reports using the same dataset. The slight improvement might stem from the experimental settings (e.g., training test set split, re-sampling techniques). Furthermore, in the domain knowledge infused feature setting we are using only 22 features out of 78 total, where each feature indicates the associated compromises (e.g., confidentiality, integrity, or availability), capable of producing better explainable and interpretable results compared to the counterpart. The prediction for a particular sample can be represented as: 
\begin{equation} \label{eq:formula6}
 P(D) = b + \sum_{g=0}^{G} contribution (g)
\end{equation}
where b is the model average and g is the generalized domain feature (e.g.,  ACK Flag Count - C), P(D) is the probability value of the decision. Instead of using contributions from each of the domain features, we can express the output in terms of the contribution from each element of the
domain concept. For that, we need to aggregate contributions from all features into three groups (C, I, and A). This will enable an analyst to understand the nature of the attack more quickly (Figure \ref{fig:breakdown}). For instance, when the greater portion of a feature contribution for a sample is from features tagged with -A (i.e., \textit{Availability}) then it might be a DDoS attack, which usually comes with very high compromises in \textit{availability} of data or service. We use the \textit{iml} package from the programming language \textit{R} to generate the breakdown of feature contributions of a particular sample's prediction (Figure \ref{fig:breakdown}).

\begin{table}
\tiny
\centering
\caption{Performance using all features vs domain features}
\label{tab:all_vs_dom}
\begin{tabular}{llllll}
\toprule
Alg.   & Acc.     & Prec.    & Rec.     & F-score  & AUC     \\
\midrule
RF-A   & \textbf{0.9987}   & \textbf{0.9965}   & \textbf{0.9971}   & \textbf{0.9968}   & \textbf{0.9997}  \\
RF-D1  & 0.9973   & 0.9920   & 0.9945   & 0.9932   & 0.9993  \\
\textit{Difference} & 0.0014   & 0.0045   & 0.0027   & 0.0036   & 0.0004  \\
ET-A   & 0.9981   & 0.9951   & 0.9951   & 0.9951   & 0.9994  \\
ET-D1  & 0.9969   & 0.9913   & 0.9932   & 0.9923   & 0.9989  \\
\textit{Difference} & 0.0011   & 0.0038   & 0.0020   & 0.0029   & 0.0004  \\
ANN-A  & 0.9802   & 0.9155   & 0.9908   & 0.9516   & 0.9984  \\
ANN-D1 & 0.9497   & 0.8300   & 0.9362   & 0.8799   & 0.9865  \\
\textit{Difference} & 0.0305   & 0.0855   & 0.0546   & 0.0717   & 0.0119  \\
SVM-A  & 0.9109   & 0.6996   & 0.9595   & 0.8092   & 0.9780  \\
SVM-D1 & 0.8489   & 0.5747   & 0.8968   & 0.7005   & 0.9252  \\
\textit{Difference} & 0.0619   & 0.1249   & 0.0627   & 0.1087   & 0.0528  \\
GB-A   & 0.9960   & 0.9854   & 0.9944   & 0.9899   & 0.9995  \\
GB-D1  & 0.9881   & 0.9513   & 0.9904   & 0.9705   & 0.9986  \\
\textit{Difference} & 0.0079   & 0.0341   & 0.0039   & 0.0194   & 0.0009  \\
NB-A   & 0.7753   & 0.4371   & 0.4888   & 0.4615   & 0.8601  \\
NB-D1  & 0.7982   & 0.4881   & 0.5028   & 0.4953   & 0.8553  \\
\textit{Difference} & (0.0229) & (0.0510) & (0.0140) & (0.0338) & 0.0048  \\
\bottomrule
\end{tabular}
\end{table}

\subsection{Findings from Generalizability Test}\label{subsec:res_generalizability}
\begin{figure}[h]
  \centering
  \includegraphics[width=\linewidth]{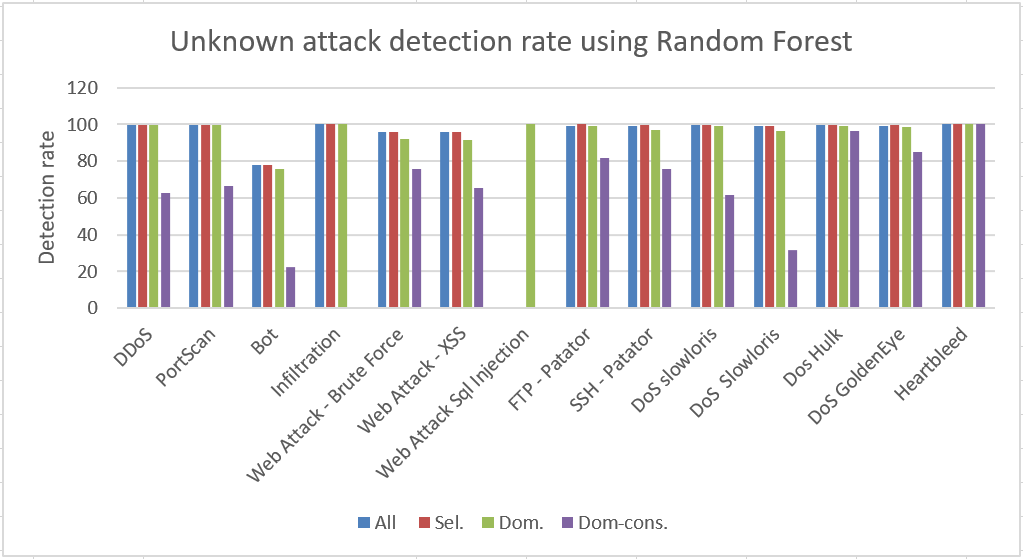}
  \caption{Unknown attack detection rate using Random Forest}
    \label{fig:detection_rate_rf}
\end{figure}

Recall that the purpose of this test is to test the resiliency against unknown attacks.
First, we use Random Forest (RF), the best performing algorithm so far, using all four settings of features. As shown in Table \ref{tab:unseen_rf} and Figure \ref{fig:detection_rate_rf}, we see that except for the constructed feature settings (denoted by Cons.), the performances of other feature settings (all, selected, and domain) are similar. The constructed features fail to provide comparable performance for RF as it has only three features and loses data details (i.e., too much generalization). Surprisingly, a few unknown attacks are only detectable using the domain knowledge infused features. For instance, \textit{Web Attack Sql Injection} is detected as suspicious only by domain knowledge infused features. Overall, although the \textit{domain knowledge infused} feature setting perform slightly worse than the \textit{all feature} setting, it comes with an explainable features set with the added capability of identifying a few unknown attacks. 

\begin{table}
\tiny
\centering
\caption{Performance of unseen attack detection using RF}
\label{tab:unseen_rf}
\begin{tabular}{llllll}
\toprule
Attack                   & Count  & All(\%) & Sel.(\%) & Dom.(\%) & Cons.(\%)  \\
\midrule
Ddos                     & 4184     & 99.90    & 99.90     & 99.90     & 62.86          \\
PortScan                 & 4973     & 99.90    & 99.94     & 99.94     & 66.28          \\
Bot                      & 54       & 77.78    & 77.78     & 75.93     & 22.22          \\
Infiltration             & 1        & 100   & 100    & 100    & 0.00           \\
Web Attack-BF            & 49       & 95.92    & 95.92     & 91.84     & 75.51          \\
Web Attack-XSS           & 23       & 95.65    & 95.65     & 91.30     & 65.22          \\
Web Attack-Sql           & 1       & \textbf{0.00}     & \textbf{0.00}      & \textbf{100}    & \textbf{0.00}           \\
FTP-Patator              & 251      & 99.20    & 100    & 99.20     & 81.67          \\
SSH-Patator              & 198      & 98.99    & 99.49     & 96.97     & 75.76          \\
DoS slowloris            & 188      & 99.47    & 99.47     & 98.94     & 61.70          \\
DoS Slowloris            & 174      & 99.43    & 99.43     & 96.55     & 31.61          \\
Dos Hulk                 & 7319     & 99.71    & 99.73     & 99.34     & 96.19          \\
DoS GoldenEye            & 314      & 99.36    & 99.68     & 98.41     & 85.03          \\
Heartbleed               & 1        & 100   & 100    & 100    & 100        \\
\bottomrule
\end{tabular}
\end{table}

To reiterate, the constructed features set consists of only three features (C, I, and A) constructed from aggregating domain knowledge infused features. As this feature setting is composed of only three features, it is an extreme generalization of features and it loses a lot of details of data. However, this time it comes with an exceptional capability which we realize after applying a statistical approach (Naive Bayes) on the dataset. We find that (see Table \ref{tab:unseen_nb}), for NB, the newly constructed feature setting is best as NB is also able to detect unknown attacks with similar accuracy compared to other feature settings by RF in Table \ref{tab:unseen_rf}. The most interesting thing about this capability is that this feature set is composed of only three features (C, I, and A), takes comparatively less time to execute, and comes with the added benefit of very good explainability. Once the prediction is expressed as a percentage of influence from each of C, I, and A, the analyst would be able to perceive the level of compromise more intuitively from the hints about the type of attack (e.g., DDoS will show a high percentage of A\textemdash{compromise in Availability}).

However, from Table \ref{tab:all_vs_sel}, \ref{tab:dom_vs_con}, and \ref{tab:all_vs_dom}, we can see that NB's performance comes at a cost of precision and recall (i.e., produces comparatively more false positives and false negatives). In addition, NB is a bad probability estimator of the predicted output  \cite{zhang2004optimality}. However, NB with constructed features setting could be recommended as an additional IDS for quick interpretation of huge traffic data given the decision is treated as tentative with the requirement of a further sanity check. We also calculate the average time taken by each algorithm for all four feature settings and found that NB is the fastest algorithm. RF, ET, GB, ANN, and SVM take 2.80, 9.27, 77.06, 15.07, and 444.50 times more execution time compared to NB. Besides, the best algorithm, RF (1\textsuperscript{st} in terms of the performance metric and 2\textsuperscript{nd} in terms of execution time), can be executed in parallel using an \textit{Apache Spark} for a far better run-time \cite{chen2016parallel} making it highly scalable to big data problems.
\begin{table}
\tiny
\centering
\caption{Performance of unseen attack detection using NB}
\label{tab:unseen_nb}
\begin{tabular}{llllll}
\toprule
Attack                   & Count  & All(\%) & Sel.(\%) & Dom.(\%) & Cons.(\%)  \\
\midrule
Ddos                     & 4184     & 76.94    & 74.59     & 83.22     & \textbf{100}        \\
PortScan                 & 4973     & 0.18     & 0.18      & 6.64      & \textbf{100}         \\
Bot                      & 54       & 0.00     &           & 0.00      & \textbf{100}         \\
Infiltration             & 1        & 0.00     & 0.00      & 0.00      & \textbf{100}         \\
Web Attack-BF          & 49       & 6.12     & 6.12      & 83.67     & \textbf{100}         \\
Web Attack-XSS           & 23       & 0.00     & 0.00      & 95.65     & \textbf{100}         \\
Web Attack-Sql           & 1        & \textbf{100}   & \textbf{100}    & 0.00      & \textbf{100}        \\
FTP-Patator              & 251      & 0.00     & 0.00      & 0.00      & 48.61          \\
SSH-Patator              & 198      & 0.00     & 0.00      & 0.00      & \textbf{100}         \\
DoS slowloris            & 188      & 25.53    & 30.85     & 38.30     & \textbf{66.49}          \\
DoS Slowloris            & 174      & 71.26    & \textbf{79.89}     & 75.29     & 78.16          \\
Dos Hulk                 & 7319     & \textbf{67.78}    & \textbf{67.78}     & \textbf{67.03}     & 35.37          \\
DoS GoldenEye            & 314      & 50.32    & 65.29     & 47.13     & \textbf{96.82}          \\
Heartbleed               & 1        & \textbf{100}   & \textbf{100}    & 0.00      & 0.00          \\
\bottomrule
\end{tabular}
\end{table}

\begin{figure}[h]
  \centering
  \includegraphics[width=\linewidth]{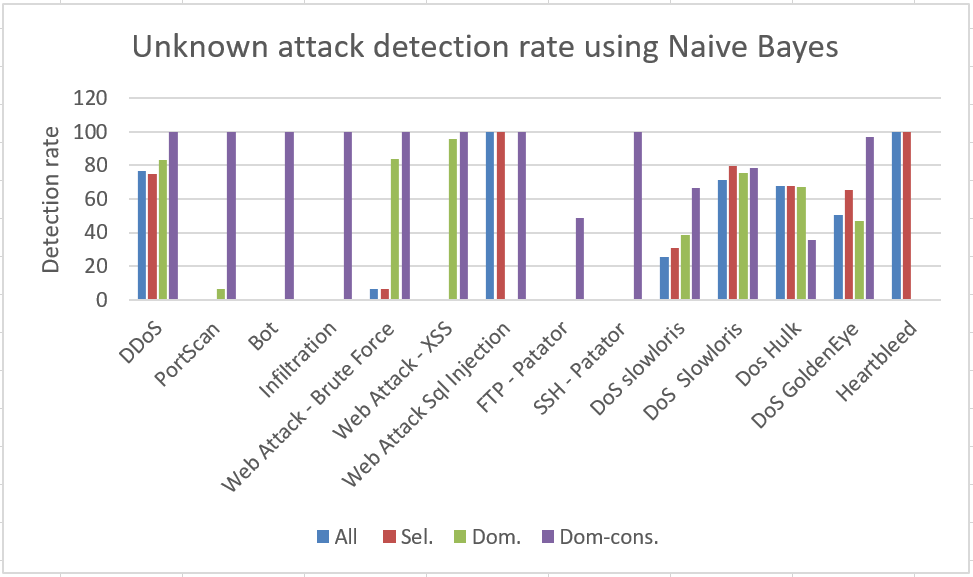}
\caption{Unknown attack detection rate using Naive Bayes}
    \label{fig:detection_rate_nb}
\end{figure}
%//
Overall, domain knowledge infusion provides better explainability with negligible compromises in performance. In addition, the generalization provides better execution time and resiliency with unknown attacks. 
%
%
%
%###################################  CONCLUSION AND FUTURE WORK ############################ 
%
%
%
\section{Conclusion and Future Work}\label{sec:conclusion}

AI-based approaches have become an integral part of security solutions due to the potential for handling ``Big Data'' and handling diverse network traffic data. Cybercrime-related damages continue to rise, and network intrusions are a key tactic. Although AI-based IDS provides accelerated speeds in intrusion \textit{detection}, \textit{response} is still at a human speed where there is human in the loop. The lack of explainability of an AI-based model is a key reason for this bottleneck. To mitigate this problem, we infuse the CIA principle (i.e., domain knowledge) in the AI-based black box model for better explainability and generalizability of the model. Our experimental results show realizable successes in better explainability with a comprehensive, up to date, and real-world network intrusion dataset.  In addition, the infused domain knowledge helps in detecting an unknown attack as it generalizes the problem, which ultimately opens the door to accommodate big data. 

Going forward, finding an optimal solution to segregate the contribution of each participating feature (sample wise) considering interactions (i.e., correlations among features complicate explanations) among features will aid in better explainability of an individual prediction (i.e., per sample). Besides, to ensure trust, estimating the level of uncertainty in the model will be another extension of this work. There are some open challenges surrounding explainability and interpretability such as an agreement of what an explanation is and to whom, a formalism for the explanation, and quantifying the human comprehensibility of the explanation.
\section*{Acknowledgment}

Thanks to Tennessee Tech's Cyber-security Education, Research and Outreach Center (CEROC) for supporting this research. 
% References and End of Paper
% These lines must be placed at the end of your paper

%\begin{flushleft}
\bibliography{bib.bib}
%\end{flushleft}
\bibliographystyle{aaai}
\end{document}